\documentclass{eptcs}
\usepackage{mdframed}
\usepackage{microtype}
\usepackage[all=normal,floats=tight,mathspacing=tight]{savetrees}
\usepackage[T1]{fontenc}
\usepackage{color}
\usepackage{cite}
\usepackage{amsmath}
\usepackage{amssymb}
\usepackage{amsfonts}
\usepackage{algorithmic}
\usepackage{textcomp}
\usepackage{xcolor}
\usepackage[export]{adjustbox}
\usepackage{subfigure}
\usepackage{graphicx}
\usepackage{booktabs} % this is for making tables.
\usepackage{commath}
\usepackage{mathtools}
\usepackage{listings}
\usepackage{ifthen}
\usepackage{xurl}
\usepackage{float}
\usepackage{xstring}
\usepackage{tikz}
\usepackage{tikz-qtree}
\usepackage{xfp}
\usepackage{caption}
\usepackage{bookmark}
\usepackage{newfloat} %needed for floating environment; grammar

\newboolean{IncludeAppendix}
\setboolean{IncludeAppendix}{true}
\newcommand{\Invalid}{\mathit{I}}
\newcommand{\Failure}{\mathit{F}}
\newcommand{\Running}{\mathit{R}}
\newcommand{\Success}{\mathit{S}}

\definecolor{BehaviorTreeBlackboardColor}{HTML}{0000FF} %blue
\definecolor{BehaviorTreeEnvironmentColor}{HTML}{FF8C00} %dark orange

\newcommand{\CiteAsText}[1]{\cite{#1}}
\newcommand{\CiteAsRef}[1]{\cite{#1}}

\newcommand{\Integers}{\mathbb{Z}}
\newcommand{\SuchThat}{\text{ s.t. }}

\newcommand{\DefinedAs}{\triangleq}

\newcommand{\CurrentChapter}{Null}
\newcommand{\CurrentSection}{Null}
\newcommand{\CurrentSubsection}{Null}
\newcommand{\CurrentSubsubsection}{Null}
\newcommand{\CurrentParagraph}{Null}

\newcommand{\LabelSection}[2]{#1::#2}
\newcommand{\LabelSubsection}[3]{#1::#2::#3}

\newcommand{\LabelParagraph}[5]{#1::#2::#3::#4::#5}
\newcommand{\LabelObject}[6]{#1::#2::#3::#4::#5::#6}

\newcommand{\LabelObjectHere}[1]{\LabelObject{\CurrentChapter}{\CurrentSection}{\CurrentSubsection}{\CurrentSubsubsection}{\CurrentParagraph}{#1}}
\newcommand{\NewSection}[1]{\renewcommand{\CurrentSection}{#1}\renewcommand{\CurrentSubsection}{Null}\renewcommand{\CurrentSubsubsection}{Null}\renewcommand{\CurrentParagraph}{Null}\section{#1}\label{\LabelSection{\CurrentChapter}{#1}}\pdfbookmark[1]{#1}{\LabelSection{\CurrentChapter}{#1}}}
\newcommand{\NewSubsection}[1]{\renewcommand{\CurrentSubsection}{#1}\renewcommand{\CurrentSubsubsection}{Null}\renewcommand{\CurrentParagraph}{Null}\subsection{#1}\label{\LabelSubsection{\CurrentChapter}{\CurrentSection}{#1}}\pdfbookmark[2]{#1}{\LabelSubsection{\CurrentChapter}{\CurrentSection}{#1}}}

\newcommand{\NewParagraph}[1]{\renewcommand{\CurrentParagraph}{#1}\paragraph{#1}\label{\LabelParagraph{\CurrentChapter}{\CurrentSection}{\CurrentSubsection}{\CurrentSubsubsection}{#1}}\pdfbookmark[4]{#1}{\LabelParagraph{\CurrentChapter}{\CurrentSection}{\CurrentSubsection}{\CurrentSubsubsection}{#1}}}
\usetikzlibrary{shapes.geometric, shapes.misc, shapes.arrows, shapes.callouts, shapes.multipart, fit, positioning, calc, matrix, automata}

\definecolor{BehaviorTreeSelectorColor}{HTML}{00FFFF}
\definecolor{BehaviorTreeSequenceColor}{HTML}{FFA500}
\definecolor{BehaviorTreeParallelColor}{HTML}{FFD700}
\definecolor{BehaviorTreeLeafColor}{HTML}{C0C0C0}

\tikzset{
    Selector/.style={
        draw=black,
        fill=BehaviorTreeSelectorColor,
        text=black,
        shape=chamfered rectangle,
        minimum size=15pt,
        inner sep=0pt,
        font=\tiny
    },
    Sequence/.style={
        draw=black,
        fill=BehaviorTreeSequenceColor,
        text=black,
        shape=rectangle,
        minimum size=15pt,
        inner sep=0pt,
        font=\tiny
    },
    Parallel/.style={
        draw=black,
        fill=BehaviorTreeParallelColor,
        text=black,
        shape=trapezium,
        trapezium left angle=60,
        trapezium right angle=120,
        minimum size=15pt,
        inner sep=0pt,
        trapezium stretches body,
        font=\tiny,
        align=center
    },
    Decorator/.style={
        draw=black,
        fill=white,
        text=black,
        shape=trapezium,
        trapezium left angle=67,
        trapezium right angle=67,
        minimum size=15pt,
        inner sep=0pt,
        trapezium stretches body,
        font=\tiny,
        align=center
    },
    Action/.style={
        draw=black,
        fill=BehaviorTreeLeafColor,
        text=black,
        shape=ellipse,
        minimum size=15pt,
        inner sep=0pt,
        font=\tiny
    },
    Check/.style={
        draw=black,
        fill=BehaviorTreeLeafColor,
        text=red,
        shape=ellipse,
        minimum size=15pt,
        inner sep=0pt,
        font=\tiny
    },
    Blackboard/.style={
        draw=blue,
        fill=white,
        text=black,
        shape=rectangle,
        font=\tiny,
        inner sep = .5pt
    },
  TreeTable/.style={
    matrix of nodes,
    row sep=0pt,
    column sep=0pt,
    nodes={
      rectangle,
      draw=black,
      align=center
    },
    minimum height=1pt,
    text depth=0.5ex,
    text height=1.5ex,
    nodes in empty cells,
    column 1/.style={
      nodes={text width=5.0em,font=\bfseries}
    }
  },
  TreeTableWide/.style={
    matrix of nodes,
    row sep=0pt,
    column sep=0pt,
    nodes={
      rectangle,
      draw=black,
      align=center
    },
    minimum width=7pt,
    minimum height=1pt,
    text depth=0.5ex,
    text height=1.5ex,
    nodes in empty cells,
    column 1/.style={
      nodes={text width=5.0em,font=\bfseries}
    }
  }
}
\newcommand{\BehaviorTree}{\mathit{BT}}

\newcommand{\Tool}{\ifthenelse{\boolean{Anon}}{TOOLNAME}{BehaVerify}}
\definecolor{pgreen}{rgb}{0.0, 0.5, 0.0}

\newcommand{\BuchiAutomaton}{BA}
\newcommand{\BuchiStates}{Q}
\newcommand{\BuchiAlphabet}{\Sigma}
\newcommand{\BuchiRelation}{\Delta}
\newcommand{\BuchiInitial}{q_0}
\newcommand{\BuchiAccepting}{F}

\newcommand{\TreeStates}{S}
\newcommand{\TreeVariables}{V}
\newcommand{\TreeMonitor}{M}
\newcommand{\TreeAlphabet}{\Sigma_{T}}
\newcommand{\TreeTransition}{\Delta_{T}}
\newcommand{\TreeTransitionMonitor}{\Delta_{M}}
\newcommand{\TreeStateInitial}{s_{0}}
\newcommand{\TreeVariableInitial}{v_{0}}
\newcommand{\TreeMonitorInitial}{m_{0}}
\newcommand{\BehaviorTreeMonitor}{BTM}

\newcommand{\LTLFormula}{\phi}
\newcommand{\LTLFormulaA}{\phi_1}
\newcommand{\LTLFormulaB}{\phi_2}

\newcommand{\LTLNext}{\bigcirc}
\newcommand{\LTLUntil}{\mathcal{U}}

\newcommand{\LTLGlobally}{\square}
\newcommand{\LTLFinally}{\lozenge}
\newcommand{\LTLSafety}{\phi_S}
\newcommand{\LTLLiveness}{\phi_L}
\newcommand{\LTLSafetyA}{\phi_{S1}}
\newcommand{\LTLLivenessA}{\phi_{L1}}

\newcommand{\LTLTrace}{Tr}
\newcommand{\LTLState}[1]{n_{#1}}

\usetikzlibrary{positioning, arrows.meta} %used by Pipeline.tex. Move somewhere else.

\definecolor{DSLCommentColor}{HTML}{778899}
\newcommand{\DSLComment}[2]{\hspace{#1}\textcolor{DSLCommentColor}{\text{\##2}}}
\DeclareFloatingEnvironment[
  fileext   = loc,
  listname  = List of Grammars,
  name      = Grammar,
  placement = htbp,
]{Grammar}

\renewcommand{\CurrentChapter}{Verification of Behavior Trees with Contingency Monitors}
\renewcommand{\CurrentSection}{Null}
\renewcommand{\CurrentSubsection}{Null}
\renewcommand{\CurrentSubsubsection}{Null}
\renewcommand{\CurrentParagraph}{Null}

\begin{document}
\title{Verification of Behavior Trees with Contingency Monitors}

\author{Serena S. Serbinowska
  \institute{{0000{-}0002{-}9259{-}1586}\\Vanderbilt University\\ Nashville TN, USA}
  \email{serena.serbinowska@vanderbilt.edu}
  \and
  Nicholas Potteiger
  \institute{{0009{-}0005{-}0406{-}0355}\\Vanderbilt University\\ Nashville TN, USA}
  \email{nicholas.potteiger@vanderbilt.edu}
  \and
  Anne M. Tumlin
  \institute{{0009{-}0000{-}1635{-}8793}\\Vanderbilt University\\ Nashville TN, USA}
  \email{anne.m.tumlin@vanderbilt.edu}
  \and
  Taylor T. Johnson
  \institute{{0000{-}0001{-}8021{-}9923}\\Vanderbilt University\\ Nashville TN, USA}
  \email{taylor.johnson@vanderbilt.edu}
}

\def\titlerunning{Verification of BT with Monitors}
\def\authorrunning{S. Serbinowska, N. Potteiger, A. Tumlin, T. Johnson}

\maketitle
\vspace{-20pt}
\begin{abstract}
  Behavior Trees ($\BehaviorTree$s) are high level controllers that have found use in a wide range of robotics tasks\@.
As they grow in popularity and usage, it is crucial to ensure that the appropriate tools and methods are available for ensuring they work as intended\@.
To that end, we created a new methodology by which to create Runtime Monitors for $\BehaviorTree$s\@.
These monitors can be used by the $\BehaviorTree$ to correct when undesirable behavior is detected and are capable of handling LTL specifications\@.
We demonstrate that in terms of runtime, the generated monitors are on par with monitors generated by existing tools and highlight certain features that make our method more desirable in various situations\@.
We note that our method allows for our monitors to be swapped out with alternate monitors with fairly minimal user effort\@.
Finally, our method ties in with our existing tool, BehaVerify, allowing for the verification of $\BehaviorTree$s with monitors\@.

\end{abstract}

\NewSection{Introduction}
A Behavior Tree ($\BehaviorTree$) is a high-level tree-structured controller with leaf nodes that interact with the environment and interior nodes that control which branches of the tree are executed\@.
The tree-structure means that $\BehaviorTree$s are often more intuitive than equivalent finite state machines, but are also powerful tools capable of being used in many environments\@.
Furthermore, the inherently recursive nature of tree structures allows for adaptability, modularity, and reuse\@.
\par
{$\BehaviorTree$}s originated in video games and were used for Non Playable Characters (NPCs)\@.
NPCs are, in essence, virtual agents in a digital environment\@.
As time progressed, NPCs needed to respond to more complex environments\@.
The video game industry responded to this by creating $\BehaviorTree$s: designer friendly controllers for complex systems\@.
In light of this, it is unsurprising that the controllers subsequently made the jump to areas such as robotics and drone control\@.
Bipedal locomotion for robots \CiteAsRef{gu2022ICRA}, vision measurement systems of road users \CiteAsRef{qin2023IEEE-Transactions-on-Cybernetics}, and swarms of agents have all utilized $\BehaviorTree$s\@.
A recent survey \CiteAsText{iovino2022Robotics-and-Autonomous-Systems} provides even more examples of $\BehaviorTree$s in action\@.
\par
It is clear that $\BehaviorTree$s are continuing to grow in popularity and usage\@.
As they expand into new domains, especially real-world safety-critical domains, it is imperative to be able to provide guarantees about their correctness\@.
Two methods for providing such guarantees are runtime monitoring and design time verification\@.
Runtime monitoring can be used to alert the $\BehaviorTree$ if there is danger of a violation occurring, allowing the tree to self-correct, while design time verification can be used to ensure the model is correct\@.
\par
At present, tools already exist for the creation of runtime monitors, such as NASA's Copilot~\CiteAsRef{perez2020NASA} and NuRV~\CiteAsRef{NuRV}, though they are not designed for $\BehaviorTree$s specifically\@.
However, it is important that the tools not only exist, but be compatible with the $\BehaviorTree$, and that the $\BehaviorTree$ reacts correctly to these tools\@.
After all, if the monitor correctly indicates a dangerous situation is occurring but the $\BehaviorTree$ ignores this warning, then the danger has not been averted\@.
\par
% To this end, we develop \Tool{}, a tool that not only generates monitors for $\BehaviorTree$s but also integrates seamlessly with the Domain Specific Language (DSL) of BehaVerify and extends BehaVerify\@.
% \Tool{} allows for the generation and verification of $\BehaviorTree$s and their associated monitors, ensuring correctness and reliability\@.
% \par
The primary contributions of this work are the following:
\begin{enumerate}
\item
  We provide a formal definition for $\BehaviorTree$s with Monitors ($\BehaviorTreeMonitor$)\@.
\item
  We expand the Domain Specific Language (DSL) of BehaVerify~\CiteAsRef{serena2022SEFM}, allowing it to describe $\BehaviorTreeMonitor$s\@.
  BehaVerify was originally created for design time verification on $\BehaviorTree$s\@. 
\item
  We present software for converting Linear Temporal Logic (LTL) specifications written in the DSL into input for the existing tool LTL2BA \CiteAsRef{gastin2001CAV}\@.
  \begin{enumerate}
  \item
    We then translate the output of LTL2BA back into the DSL\@.
    This enables BehaVerify to generate nuXmv~\CiteAsRef{nuXmv} models, allowing us to use Design Time Verification to confirm that the $\BehaviorTreeMonitor$ works as intended\@.
  \item
    Additionally, we can translate the output of LTL2BA directly into a C or Python monitor, for use with code generated by BehaVerify\@.
  \item
    Furthermore, we compare the generated C monitors to monitors generated by Copilot \CiteAsRef{perez2020NASA} and demonstrate our monitors are on par in terms of performance and offer certain improvements in terms of correctness\@.
  \end{enumerate}
\end{enumerate}

\NewSection{Related Work} % ---------------------------------------------------------------------- Section :: Related Work and Contributions
There is a broad body of work utilizing behavior trees for planning purposes in robotic systems~\cite{ogren2022annurev,colledanchise2014IEEE-RSJ,biggar2021IEEE-Robotics-and-Automation-Letters,colledanchise2017IEEE,ogren2020IEEE-Robotics-and-Automation-Letters,sprague2022IEEE,colledanchise2016IROS,marzinotto2014IEEE}, illustrating their broad usage in safety-critical systems\@.
There are several practical implementations of $\BehaviorTree$s (such as PyTrees \CiteAsRef{PyTrees} and its Robotic Operating System (ROS) extension PyTreesRos, BehaviorTree.cpp \CiteAsRef{BehaviorTreeCPP}, and Unreal Engine \CiteAsRef{UnrealEngine})\@.
Each of these feature a Blackboard (shared memory between nodes)\@.
For a variety of practical reasons, our tool targets the implementation of $\BehaviorTree$s presented in PyTrees, though we hope to target BehaviorTree.cpp in the future as well\@.
\par
There are several existing works that develop and apply formal verification for $\BehaviorTree$s\@.
There are several tools for model verification of $\BehaviorTree$s: \CiteAsText{biggar2020IEEE-Robotics-and-Automation-Letters}, BehaVerify~\CiteAsRef{serena2022SEFM}, BTCompiler, ArcadeBT~\CiteAsRef{henn2022FMICS}, and MoVe4BT~\CiteAsRef{MoVe4BT}\@.
Prior to this work, however, none of these tools supported Runtime Verification of $\BehaviorTree$s\@.
\CiteAsText{colledanchise2021IROS} does runtime verification for a fragment of Timed Propositional Temporal Logic (TPTL) for $\BehaviorTree$s, but failed to configure our examples to work with it\@.
\par
The authors in \CiteAsText{perez2022TACAS} utilized the runtime verification framework NASA Copilot~\CiteAsRef{perez2020NASA} to ensure that a flying aircraft maintains an airspeed above a threshold using natural language and Past Linear Temporal Logic (PLTL)\@.
Similarly, stream runtime verification (SRV) monitors were generated using HLola~\CiteAsRef{gorostiaga2021TACAS} in \CiteAsText{zudaire2021ICRA} to seamlessly integrate with a UAV hybrid navigation controller for post decision making and online remediation actions\@.
Furthermore, the authors in \CiteAsText{will2023IFIP} develop an architecture that allows for construction of runtime monitors that can be integrated into an Urban Air Mobility (UAM) System\@.
They demonstrate that runtime monitors are built using NASA OGMA \CiteAsText{perez2022TACAS} and NASA FRET \CiteAsText{perez2022TACAS} for battery monitoring of a UAV simulated in Microsoft AirSim~\CiteAsRef{shah2017FSR}\@.
Runtime monitoring instrumentation frameworks for ROS specifically also have been developed~\cite{ferrando2020TAROS}.
Our approach generally differs from these works because we are interested in creating monitors for $\BehaviorTree$s specifically\@.
However, our approach also clearly has overlap with several of these methods; we too seek to enable remediation actions for the models being monitored\@.
Furthermore, while we create our own monitors, our general framework is compatible with alternative monitors\@.
In light of this, we compare our monitors to those created by Copilot in Section~\ref{\LabelSection{\CurrentChapter}{Monitor Comparison}}\@.
\par{}
Another related aspect is that of the Simplex architecture~\cite{seto1998ACC,seto1999DTIC,bak2009RTAS}\@.
The Simplex architecture describes control switching logic swaps between multiple controllers depending on the current state of the system\@.
Our work differs from this approach in two main methods: first our approach is focused explicitly on $\BehaviorTree$s, and second our approach does not utilize multiple controllers\@.
Instead, we utilize the structure of $\BehaviorTree$s to integrate the `switch' into the $\BehaviorTree$ itself\@.
\par{}
The work that is most closely related to ours is \CiteAsText{colledanchise2021IROS}\@.
This work is about $\BehaviorTree$s equipped with runtime monitors based on Timed Propositional Temporal Logic (TPTL) and provides a formal definition of the setup\@.
However, unlike our work, the monitors are not meant to be interacted with; the $\BehaviorTree$ has no way of reacting to a violation\@.
In terms of generating monitors by transforming temporal logic specifications to automata, our approach is similar to that of ltl2mon and LamaConv~\cite{bauer2011TOSEM,elhokayem2018RV,ferrando2022PAAMS}, but differs in that we consider $\BehaviorTree$s.
We further differentiate our work through the design time verification aspect\@.
Our method allows us to prove that a $\BehaviorTree$ equipped with a monitor and its contingency response for detected violations is guaranteed to satisfy a specification\@.
% \par{}
% In contrast to these existing works, in this paper, we present a framework for utilizing monitors of safety and temporal logic properties within behavior trees\@.
% This allows us to consider both verification of specifications for the planning process in the behavior tree utilizing the contingency monitor at design-time (e.g., to verify if the monitor is triggered that eventually it executes\todo{add a reasonable example or clarify}), as well as reasoning about runtime verification of behavior utilizing the monitors\@.
% These design-time and runtime verification approaches are evaluated on several robotics scenarios, where such contingencies are commonly necessary and important for ensuring safety and liveness at runtime\@.

\NewSection{Preliminaries}
This section provides a formal definition for Linear Temporal Logic and Buchi Automata\@.
This is followed by an intuitive overview of $\BehaviorTree$s and a formal definition\@.
\NewSubsection{Linear Temporal Logic}
A Linear Temporal Logic (LTL) formula is evaluated on a trace\@.
A trace is a sequence of states $\LTLTrace \DefinedAs [\LTLState{0}, \LTLState{1}, \ldots]$\@.
Here $\LTLState{0}$ is the state at time 0, $\LTLState{1}$ is the state at time 1, etc\@.
The grammar of an LTL formula is presented in Grammar~\ref{\LabelObjectHere{LTL Grammar}}\@.
\par{}
\begin{mdframed}
  % \begin{grammar}
%   <LTL> :: = <a> \DSLComment{10pt}{First Order Logic Formula}
%   \alt \ensuremath{\neg} <LTL> | <LTL> \ensuremath{\lor} <LTL> \DSLComment{10pt}{Boolean operators; in practice we allow more operators}
%   \alt \ensuremath{\LTLNext}(<LTL>) | (<LTL>)\ensuremath{\LTLUntil}(<LTL>) \DSLComment{10pt}{Temporal operators next and until}
% \end{grammar}
\[
\begin{array}{rcll}
\langle \text{LTL} \rangle & ::= & \langle \text{a} \rangle & \DSLComment{10pt}{First Order Logic Formula} \\
                            &  |  & \neg \langle \text{LTL} \rangle \mid \langle \text{LTL} \rangle \lor \langle \text{LTL} \rangle & \DSLComment{10pt}{Minimal Boolean operators} \\
                            &  |  & \LTLNext (\langle \text{LTL} \rangle) \mid (\langle \text{LTL} \rangle) \LTLUntil (\langle \text{LTL} \rangle) & \DSLComment{10pt}{Temporal operators next and until}
\end{array}
\]
\captionsetup{hypcap=false}
\captionof{Grammar}{Minimal LTL Grammar.}\label{\LabelObjectHere{LTL Grammar}}
\captionsetup{hypcap=true}

\end{mdframed}
% \begin{grammar}[Minimal LTL Grammar][h][\LabelObjectHere{LTL Grammar}]
%   \firstcase{\LTLFormula}{\LTLAtom}{First Order Logic Formula}
%   \otherform{\neg \LTLFormula \gralt \LTLFormulaA \lor \LTLFormulaB}{Boolean operators on LTL formulas}
%   \otherform{\LTLNext{\LTLFormula} \gralt \LTLUntil{\LTLFormulaA}{\LTLFormulaB}}{Temporal operators on next and until on LTL formulas}
% \end{grammar}
We assume the reader is familiar with Boolean logic, so we will not describe them here\@.
$\LTLNext(\LTLFormula)$ is true at time $t$ if $\LTLFormula$ is true at time $t+1$\@.
$\LTLFormulaA\LTLUntil\LTLFormulaB$ is true at time $t$ if $\exists t^{\prime\prime}$ such that $t \leq t^{\prime\prime}$ and $\LTLFormulaB$ is true at $t^{\prime\prime}$ and $\forall t^{\prime}$ such that $t \leq t^{\prime} < t^{\prime\prime}$, $\LTLFormulaA$ is true at $t^{\prime}$\@.
\par{}
In addition to the grammar presented in Grammar~\ref{\LabelObjectHere{LTL Grammar}}, we also utilize $\LTLGlobally(\LTLFormula)$ (globally) and $\LTLFinally(\LTLFormula)$ (finally)\@.
These do not increase the expressiveness of LTL, but make writing formulas easier\@.
$\LTLGlobally(\LTLFormula)$ is true at time $t$ if $\forall t^{\prime}$ such that $t \leq t^{\prime}$, $\LTLFormula$ is true at time $t^{\prime}$\@.
$\LTLFinally(\LTLFormula)$ is true at time $t$ if $\exists t^{\prime}$ such that $t \leq t^{\prime}$, $\LTLFormula$ is true at time $t^{\prime}$\@.
\par{}
Finally, we say that $\LTLFormula$ is true for the entire trace if it is true at time 0\@.
Notationally, we will write $Tr \vDash \LTLFormula$ to mean that $\LTLFormula$ is true for the entire trace $Tr$, $Tr_{[i, j]} \vDash \LTLFormula$ if we are looking at the segment of the trace $[\LTLState{i}, \LTLState{i + 1}, \ldots \LTLState{j}]$, and utilize $\not\vDash$ in the same way but to mean not true\@.
\NewSubsection{Buchi Automata}
A Buchi Automaton ($\BuchiAutomaton$) is a tuple $(\BuchiStates, \BuchiAlphabet, \BuchiRelation, \BuchiInitial, \BuchiAccepting)$\@.
\begin{enumerate}
\item
  $\BuchiStates$ is a finite set representing the states $\BuchiAutomaton$ can be in\@.
\item
  $\BuchiAlphabet$ is a finite set representing the possible inputs\@.
\item
  $\BuchiRelation$ is a function from $\BuchiStates \times \BuchiAlphabet \mapsto 2^{\BuchiStates}$\@.
  Here $2^{\BuchiStates}$ is the power set of $\BuchiStates$\@.
  This function describes the nondeterministic transitions available from a given state-input combination\@.
\item
  $\BuchiInitial$ is an element of $\BuchiStates$\@.
  This is the initial state\@.
\item
  $\BuchiAccepting$ is a finite set such that $\BuchiAccepting \subseteq \BuchiStates$\@.
  This is the set of accepting states\@.
\end{enumerate}
Then, $\BuchiAutomaton$ accepts a given sequence of inputs $[a_0, a_1, \ldots]$ if and only if there exists a sequence $[q_0, q_1, \ldots]$ such that
\begin{enumerate}
\item
  $\forall j \in \Integers \SuchThat j \geq 0, q_{j+1} \in \BuchiRelation(q_{j}, a_{j})$
\item
  $\forall j \in \Integers, \exists k \in \Integers \SuchThat j < k \land a_{k} \in \BuchiAccepting$
\end{enumerate}
Thus we accept if we begin in the initial state, take valid transitions, and enter accepting states an infinite number of times\@.
We accept if any such trace is possible; thus a single trace \textit{can} be used to prove that the input is accepted but it \textit{cannot} be used to prove that the input is \textit{not} accepted\@.
\NewSubsection{Behavior Tree Overview}
$\BehaviorTree$s are rooted trees with parent-child relationships\@.
Each node has one parent, except the root which has no parent\@.
When executing, a $\BehaviorTree$ starts from the root and follows a Depth First Traversal\@.
Nodes can change this traversal order and leave certain branches unexplored based on what their children return\@.
This process is started by an external activation signal called a \textit{tick}\@.
In practice, trees are often structured recursively and parents propagate this signal to their children, but for this paper a \textit{tick} will only be used to refer to the root receiving the external signal\@.
Note that $\BehaviorTree$s are inactive until they receive a tick\@.
In the interest of conciseness, we will omit these periods of inactivity from various diagrams\@.
We assume that a tick will only arrive while the tree in inactive\@.
\par
Each node is always in one of four states: Success ($\Success$), Failure ($\Failure$), Running ($\Running$), or Invalid ($\Invalid$)\@.
We will use \textit{active} and \textit{executing} to describe where we are in the execution of the tree\@.
When a new tick arrives, each node is set to $\Invalid$ and the root becomes both active and executing\@.
Until the root finishes executing, exactly one node will be active at all times but more than one node can be executing\@.
A node that is executing is similar to a function that has been called but has not yet turned\@.
A node that is active is similar to a function that is currently being stepped through\@.
When a node finishes executing it returns $\Success$, $\Running$, or $\Failure$\@.
\par
In addition to ticks, we use \textit{timestep} or $t$ to track each time the active node changes\@.
Both tick and timestep will be enumerated sequentially starting from 1\@.
Refer to Figure~\ref{\LabelObject{\CurrentChapter}{\CurrentSection}{\CurrentSubsection}{Null}{Composite Nodes}{Sequence Execution Example}} for an example\@.
\par
Nodes can be grouped into three categories: leaf, decorator, and composite\@.
\NewParagraph{Leaf Nodes}
Leaf nodes do not have children\@.
It is common to categorize leaf nodes as checks/guards (\textit{e.g.} at boundary?) and actions (\textit{e.g.} go forward)\@.
Checks evaluate a boolean condition and return $\Success$ if true and $\Failure$ otherwise\@.
Consider Subfigure (a) of Figure~\ref{\LabelObjectHere{Leaf Nodes Example}}; if there is an apple on the table, the check will return $\Success$\@.
If there is not, $\Failure$ will be returned\@.
Either way, the status will be returned to the root which will proceed accordingly\@.
It is important to note that checks \textit{only} check a condition and return the appropriate status; they do not set variables or take any sort of action.\@.
By contrast, actions can execute actions, for lack of a better word\@.
For instance, in Subfigure (b) of Figure~\ref{\LabelObjectHere{Leaf Nodes Example}}, the action executes the action of moving left\@.
It is important to note that actions are also not restricted in what status they return\@.
While Subfigure (b) of Figure~\ref{\LabelObjectHere{Leaf Nodes Example}} shows $\Failure$ being returned, it would be valid to create a version of this action that always return $\Success$, or it could return $\Running$ because it hasn't finished, or it could return $\Failure$ based on some sort of conditional logic\@.
\begin{figure}
  \centering
  \begin{adjustbox}{max width = .9\linewidth}
    \usetikzlibrary{fit}
\usetikzlibrary{positioning, calc}
% \usetikzlibrary{calc}

\begin{tikzpicture}
  %%% Leaf nodes
  \node[Check](chk) at (8, 0) {Check};
  \node[Blackboard, align=left](chkCode) at (8, -1.0) {\normalsize Is the apple on\\ \normalsize the table?};
  \node[Sequence](root) at (8, 1.25) {Root};

    \draw[-] (chk) -- (chkCode);
    \draw[-] (chk) -- (root);
    \draw[->, dashed] ($(chk) + (-0.2, 0.9)$) -- ($(chk) + (-0.2, .3)$);
    \draw[->, dashed] ($(chk) + (0.2, .3)$) -- ($(chk) + (0.2, 0.9)$);
    \node[] at (7.2, 0.7) {tick};    
    \node[] at (9.2, 0.7) {Return=S};  
    \draw[color=red] (7.2, 0.3) circle [radius=0.2] node {1};
    \draw[color=red] (9.2, 0.3) circle [radius=0.2] node {2};

  \node[Action](act) at (12, 0) {Action};
  \node[Blackboard](actCode) at (12, -1.0) {\normalsize Move left};
  \node[Sequence](root2) at (12, 1.25) {Root};

  \draw[-] (act) -- (root2);
  \draw[-] (act) -- (actCode);
    \draw[->, dashed] ($(act) + (-0.2, 0.9)$) -- ($(act) + (-0.2, .3)$);
    \draw[->, dashed] ($(act) + (0.2, .3)$) -- ($(act) + (0.2, 0.9)$);
    \node[] at (11.2, 0.7) {tick};    
    \node[] at (13.2, 0.7) {Return=F};  
    \draw[color=red] (11.2, 0.3) circle [radius=0.2] node {1};
    \draw[color=red] (13.2, 0.3) circle [radius=0.2] node {2};

  \node[Check](chk3) at (16, 0) {Check};
  \node[Blackboard, align=left](chkCode3) at (16, -1.0) {\normalsize Is the apple on\\ \normalsize the table?};
  \node[Sequence](root3) at (16, 2.0) {Root};
  \node[Decorator, minimum width=2cm](inv) at (16, 1.0) {Invert};

    \draw[-] (chk3) -- (chkCode3);
    \draw[-] (chk3) -- (inv);
    \draw[-] (inv) -- (root3);
    \draw[->, dashed] ($(chk3) + (-0.2, 0.7)$) -- ($(chk3) + (-0.2, .3)$);
    \draw[->, dashed] ($(chk3) + (0.2, .3)$) -- ($(chk3) + (0.2, 0.7)$);
    \node[] at (15.2, 0.5) {tick};   
    \draw[color=red] (14.6, 0.5) circle [radius=0.2] node {1};
    \node[] at (17.2, 0.5) {Return=S};   
    \draw[color=red] (18.4, 0.5) circle [radius=0.2] node {2};
    \draw[->, dashed] ($(inv) + (0.2, .3)$) -- ($(inv) + (0.2, 0.7)$);
     \node[] at (17.2, 1.6) {Return=F};
     \draw[color=red] (18.4, 1.6) circle [radius=0.2] node (lbl3) {3};
     
    \node[right=0.25 of lbl3, align=center] (rtrn3) {\textcolor{red}{Return}}; 
    \node[below=0.001 of rtrn3, align=center] (rtrn4) {\textcolor{red}{inverted}}; 
    \draw[->, red] (lbl3) -- (rtrn3);

    \node[below=0.3 of chkCode, align=center] (chkCodeLabel) {(a) Check Node (\textbf{Leaf})}; 
    \node[right=0.5 of chkCodeLabel, align=center] (chkCodeLabel2) {(b) Action Node (\textbf{Leaf})}; 
    \node[right=0.5 of chkCodeLabel2, align=center] (chkCodeLabel3) {(c) Inverter Node (\textbf{Decorator})}; 
    
\end{tikzpicture}
  \end{adjustbox}
  \caption{
    Example Leaf and Decorator Nodes\@.
  }\label{\LabelObjectHere{Leaf Nodes Example}}
  %\vspace{-1em}
\end{figure}
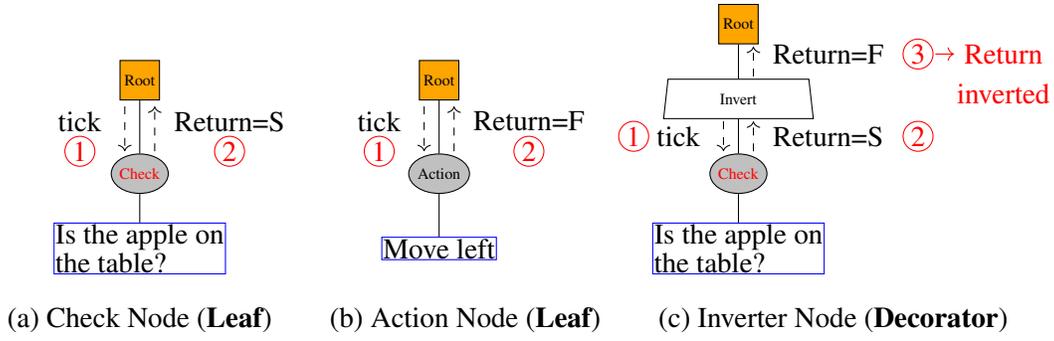%
\NewParagraph{Decorator Nodes}
Decorator nodes `decorate' their children, allowing for easy adjustments to be made\@.
A decorator node will always have exactly one child, but that child can have children of its own\@.
An inverter (decorator that swaps $\Success$ and $\Failure$) is in Subfigure (c) of Figure~\ref{\LabelObject{\CurrentChapter}{\CurrentSection}{\CurrentSubsection}{Null}{Leaf Nodes}{Leaf Nodes Example}}\@.
\NewParagraph{Composite Nodes}
Composite nodes control the traversal of the tree\@.
Changing a composite node will change the conditions under which branches of the tree are activated\@.
The three primary types of composite nodes are selector, sequence, and parallel nodes\@.
The children of composite nodes are ordered and are activated according to the order, which we will treat as being left-to-right, both visually and in our language\@.
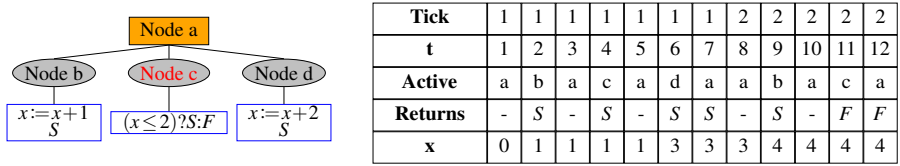
\begin{figure}
  \centering
  \begin{adjustbox}{max width = 0.75\linewidth}
    \begin{tikzpicture}
  \tikzset{level distance=25pt}
  \tikzset{sibling distance=5pt}

  \Tree [.\node[Sequence] (a){\begin{tabular}{c}\normalsize Node a\end{tabular}};
  [.\node[Action] (b) {\normalsize Node b}; \node[Blackboard]{\begin{tabular}{c}\normalsize $x \coloneqq x + 1$\\\normalsize $\Success$\end{tabular}};]
  [.\node[Check] (c) {\normalsize Node c}; \node[Blackboard]{\begin{tabular}{c}\normalsize $(x \leq 2)?\Success${:}$\Failure$\end{tabular}};]
  [.\node[Action] (d) {\normalsize Node d}; \node[Blackboard]{\begin{tabular}{c}\normalsize $x \coloneqq x + 2$\\\normalsize $\Success$\end{tabular}};]
  ]

  \matrix (table) [TreeTable, text width={1em}, right={3.0cm of a}, yshift={-1cm}]
  {
    Tick      & 1 & 1 & 1 & 1 & 1 & 1 & 1 & 2 & 2 & 2 & 2 & 2\\
    t         & 1 & 2 & 3 & 4 & 5 & 6  & 7 & 8 & 9 & 10 & 11 & 12\\
    Active    & a & b & a & c & a & d & a & a & b & a & c & a\\
    Returns   & {-} & $\Success$ & {-} & $\Success$ & {-} & $\Success$ & $\Success$ & {-} & $\Success$ & {-} & $\Failure$ & $\Failure$\\
    x         & 0 & 1 & 1 & 1 & 1 & 3 & 3 & 3 & 4 & 4 & 4 & 4\\
  };
\end{tikzpicture}
  \end{adjustbox}
  \caption{
    A $\BehaviorTree$ consisting of a sequence node (a) with two actions (b, d) and a check (c)\@.
    We use the ternary operator $i?j:k$ to mean if $i$ then $j$ else $k$\@.
    Tick indicates the number of times the tree has been ticked\@.
    At each timestep $t$, the variable $x$ is updated based on the active node\@.
    If a node is finished, then it returns one of $\Success$, $\Failure$, or $\Running$\@. 
  }\label{\LabelObjectHere{Sequence Execution Example}}
  %\vspace{-1.5em}
\end{figure}%
\begin{enumerate}
\item
  Recall our notation of Success ($\Success$), Running ($\Running$), and Failure ($\Failure$)\@.
\item
  \textbf{Selector Nodes:} Selector/Fallback nodes activate their children from left-to-right until one of them returns $\Success$ or $\Running$, at which point the selector itself returns $\Success$ or $\Running$\@.
  If a child node returns $\Failure$, the selector activates the next child\@.
  Selectors can be used to prioritize behaviors, ordering them from most preferable at the left to least preferable at the right\@.
  Another way to think of these nodes as providing a series of fallbacks in case the intended action fails\@.
\item
  \textbf{Sequence Nodes:} Sequence nodes activate their children in a left-to-right order, requiring each child to return $\Success$ before moving on to the next\@.
  The sequence returns $\Success$ only if all its children return $\Success$\@.
  It returns $\Failure$ or $\Running$ as soon as one of its children does\@.
  This is ideal for defining a series of actions that must be performed in a specific order to achieve a goal\@.
  Figure~\ref{\LabelObject{\CurrentChapter}{\CurrentSection}{\CurrentSubsection}{Null}{Composite Nodes}{Sequence Execution Example}} shows an example sequence node with two actions and one check\@.
  The figure clearly demonstrates the order in which the children become active\@.
  Additionally, it showcases how composite nodes control the flow of logic in a $\BehaviorTree$\@; during tick 1, all three children are active at some point\@.
  However, during tick 2 node (d) never becomes active because node (c) returned $\Failure$\@.
\item
  \textbf{Parallel Nodes:} For the purpose of this paper, we do not consider truly parallel nodes (i.e., nodes that activate all their children simultaneously)\@.
  Instead, our parallel nodes activate nodes one at a time in a left-to-right order\@.
  However, unlike selector and sequence nodes, a parallel node will \textit{always} tick all of its children\@.
  Once the last child returns, the parallel node uses a policy that considers what all of the children returned\@.
  The two most common policies are Success on All, which requires that all children return $\Success$ for $\Success$ to be returned, and Success on One, which requires only one child to return $\Success$ for $\Success$ to be returned\@.
  This is consistent with \CiteAsText{PyTrees}\@.
  It is possible to define more complex policies, but this is beyond the scope of this paper\@.
\end{enumerate}
\NewSubsection{Formal Definition of Behavior Trees}
A Behavior Tree ($\BehaviorTree$) is a tuple $(\TreeStates, \TreeVariables, \TreeAlphabet, \TreeTransition, \TreeStateInitial, \TreeVariableInitial)$\@.
\begin{enumerate}
\item
  $\TreeStates$ is a finite set that describes the state of the tree itself (which node is active, what nodes have returned so far, etc)\@.
\item
  $\TreeVariables$ is a finite set that describes the state of the Blackboard variables (the persistent memory of the tree)\@.
\item
  $\TreeAlphabet$ is a finite set of all possible `inputs' to the tree (e.g., environmental factors)\@.
\item
  $\TreeTransition$ is a function $\TreeStates \times \TreeVariables \times \TreeAlphabet \mapsto \TreeStates \times \TreeVariables$\@.
  This function takes the current state of the tree, the variables, and environmental factors and produces a new tree state and new variable values\@.
\item
  $\TreeStateInitial$ is an element of $\TreeStates$ that describes the initial state of the tree\@.
\item
  $\TreeVariableInitial$ is an element of $\TreeVariables$ that describes the initial state of the Blackboard variables\@.
\end{enumerate}
This definition does not allow for nondeterminism, though it could be simulated through a `seed' variable\@.
Furthermore, this definition is slightly too permissive; additional restrictions would need to be placed on $\TreeTransition$ to enforce the tree structure\@.
We omit these details as we believe that they would complicate the issue without providing additional insight\@.
\par
For a given sequence of inputs $[a_0, a_1, \ldots]$, the corresponding $\BehaviorTree$ trace is a sequence $[(s_0, v_0), (s_1, v_1), \ldots]$ such that
\begin{equation*}
  \forall j \in \Integers \SuchThat j \geq 0, \TreeTransition(s_j, v_j, a_j) = (s_{j+1}, v_{j+1})
\end{equation*}

\NewSection{Problem Statement and Methodology}
While we have defined $\BehaviorTree$s, we have not defined Behavior Trees with Monitors ($\BehaviorTreeMonitor$s)\@.
In this section we will first define $\BehaviorTreeMonitor$, then formally state the problem we are addressing, and finally present our method for addressing the issue\@.
\NewSubsection{Formal Definition of Behavior Trees with Monitors}
A $\BehaviorTreeMonitor$ is a tuple $(\TreeStates, \TreeVariables, \TreeMonitor, \TreeAlphabet, \TreeTransitionMonitor, \TreeStateInitial, \TreeVariableInitial, \TreeMonitorInitial)$\@.
\begin{enumerate}
\item
  $\TreeStates$, $\TreeVariables$, $\TreeAlphabet$, $\TreeStateInitial$, and $\TreeVariableInitial$ are unchanged from $\BehaviorTree$\@.
\item
  $\TreeMonitor$ is a finite set describing the state of the monitor\@.
\item
  $\TreeTransitionMonitor$ is a function $\TreeStates \times \TreeVariables \times \TreeAlphabet \times \TreeMonitor \mapsto \TreeStates \times \TreeVariables \times \TreeMonitor$\@.
  This function takes the current state of the tree, the variables, environmental factors, and the state of the monitor and produces a new tree state, new variables state, and a new monitor state\@.
\item
  $\TreeMonitorInitial$ is an element of $\TreeMonitor$ and describes the initial state of the monitor\@.
\end{enumerate}
\par
For a given sequence of inputs $[a_0, a_1, \ldots]$, the corresponding $\BehaviorTreeMonitor$ trace is a sequence $[(s_0, v_0, m_0),$ $(s_1, v_1, m_1),$ $\ldots]$ such that
\begin{equation*}
  \forall j \in \Integers \SuchThat j \geq 0, \TreeTransition(s_j, v_j, a_j, m_j) = (s_{j+1}, v_{j+1}, m_{j+1})
\end{equation*}
\NewSubsection{Problem Statement}
Our problem statement follows\@.
\textbf{
  Given a $\BehaviorTree$ and a $\LTLFormula$, create a $\BehaviorTreeMonitor$ that monitors $\LTLFormula$ such that $\BehaviorTreeMonitor$ is capable of reacting to a violation of $\LTLFormula$\@.
}
To demonstrate the utiltity of this process, we will also consider $\LTLFormulaA$, a second specification that holds for $\BehaviorTreeMonitor$ but not for $\BehaviorTree$\@.
\par{}
To accomplish this goal we modified the Domain Specific Language (DSL) of BehaVerify (outlined below) to allow the use of monitors within $\BehaviorTree$s\@.
We utilize the tool LTL2BA~\CiteAsRef{gastin2001CAV} and specifically the implementation at~\footnote{\url{https://github.com/utwente-fmt/ltl2ba}} to translate an LTL formula into a monitor in the form of a $\BuchiAutomaton$.
Then we create an implementation of that $\BuchiAutomaton$ for use with the $\BehaviorTree$\@.
Finally, to verify that $\BehaviorTreeMonitor$ satisfies the property, we also convert the output of LTL2BA into an implementation of the monitor within the DSL and utilize BehaVerify to create a nuXmv model that proves $\LTLFormulaA$ is true for $\BehaviorTreeMonitor$ but false for $\BehaviorTree$\@.
\NewSubsection{BehaVerify}
BehaVerify uses a Domain Specific Language (DSL) that allows the user to specify a $\BehaviorTree$\@.
As the DSL itself~\footnote{\url{https://github.com/verivital/behaverify/blob/main/metamodel/behaverify.tx}} is complex, we will provide an overview here\@.
\par{}
The user defines a finite set of typed variables that will be used by the $\BehaviorTree$\@.
The user also defines a finite set of leaf nodes\@.
Each leaf node is a finite sequence of statements, exactly one of which is a return statement while the rest are variable statements\@.
A variable statement consists of the variable whose value is being updated and a sequence of `if statements' that determines the new value\@.
Nondeterminism is allowed in variable statements\@.
The return statement is similar, but is used to determined what status the node will return\@.
Note that the node does \textit{not} stop execution when the return statement is executed; it is only used to determine the return status, not to `return' from the node\@.
Finally, the user creates a finite tree consisting of composite (selector, sequence, and parallel), decorator (inverter and X\_is\_Y), and user-defined leaf nodes\@.
\NewParagraph{Monitors}
We add special syntax to our DSL for the creation and use of monitors\@.
The user provides an LTL specification that is to be monitored\@.
Furthermore, the user specifies where in the tree the monitor should be used, and how the tree should react to the possible outputs of the monitor\@.
We describe the details of transforming the monitor in Subsection~\ref{\LabelSubsection{\CurrentChapter}{\CurrentSection}{Generating Implementations}}\@.
\par{}
We present two pipelines for making monitors\@.
The first is for generating a Python implementation of a $\BehaviorTree$ and either Python or C code for the monitor(s)\@.
The second pipeline is for generating a nuXmv model of a $\BehaviorTree$ and its monitor(s)\@.
\NewSubsection{Generating Implementations}
\begin{figure}
  \begin{minipage}{.6\linewidth}
    \begin{adjustbox}{max width = \linewidth}
      \begin{tikzpicture}[
    box/.style = {draw, minimum width=3cm, minimum height=1cm, align=center},
    arrow/.style = {->, thick},
    label/.style = {midway, fill=white, inner sep=1pt}
]

% Nodes
\node[box] (dsl) {DSL};
\node[box, below=of dsl] (gen_py_code) {Python Code};
\node[box, right=of dsl, xshift=70pt] (ltl2ba_cmd) {Commands for LTL2BA};
\node[box, below=of ltl2ba_cmd] (spin_mon) {Spin Never Claim};
\node[box, below=of spin_mon] (py_mon) {Python/C Monitors};
\node[box, below=of gen_py_code, xshift=0pt] (gen_py_code_mon) {Python $\BehaviorTreeMonitor$};

% Arrows
\draw[->, dashed] (dsl) -- node[label] {BehaVerify} (gen_py_code);
\draw[arrow, blue] (dsl) -- (ltl2ba_cmd);
\draw[->, dashed] (ltl2ba_cmd) -- node[label] {LTL2BA} (spin_mon);
\draw[arrow, blue] (spin_mon) -- (py_mon);
\draw[arrow, blue] (gen_py_code) -- (gen_py_code_mon);
\draw[arrow, blue] (py_mon) -- (gen_py_code_mon);

\end{tikzpicture}
    \end{adjustbox}
  \end{minipage}
  \begin{minipage}{.39\linewidth}
    \caption{
      Diagram of how BehaVerify generates Python code for $\BehaviorTree$s with monitors\@.
      LTL2BA is a tool for converting an LTL specification to a $\BuchiAutomaton$\@.
      Spin is a model checker, and a Never Claim can be checked using Spin\@.
      Solid blue arrows mark new contributions\@.
    }\label{\LabelObjectHere{Python Pipeline}}
  \end{minipage}
\end{figure}
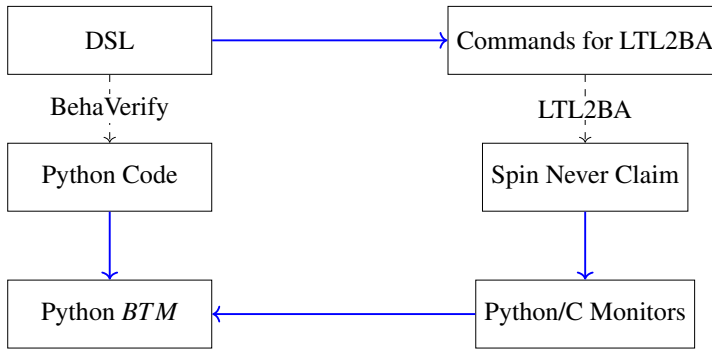
We utilize the following process to generate a $\BehaviorTreeMonitor$ using BehaVerify\@.
\begin{enumerate}
\item
  The user creates a DSL file specifying the $\BehaviorTree$ and any monitors it uses\@.
\item
  BehaVerify creates a Python implementation for $\BehaviorTree$, ignoring the monitors\@.
\item
  For each monitor in the DSL file, BehaVerify creates a command for use with LTL2BA\@.
\item
  For each command, LTL2BA creates a Buchi Automaton ($\BuchiAutomaton$)\@.
  The output is in the form of a `never claim' for use with the Spin~\CiteAsRef{SPIN} model checker\@.
\item
  For each $\BuchiAutomaton$, BehaVerify creates a corresponding Python implementation\@.
\item
  The monitors are combined with the generated Python code\@.
\end{enumerate}
This process can be seen in Figure~\ref{\LabelObjectHere{Python Pipeline}}\@.
Below we provide some additional details\@.
\NewParagraph{Command Creation}
LTL2BA commands can contain temporal operators (e.g{.} `globally'), boolean operators (e.g{.} `and'), boolean constants (`true' and `false'), and lowercase alphanumeric strings representing boolean variables\@.
Our conversion process prioritizes making the resulting LTL formula as small as possible\@.
Thus the formula $\LTLGlobally(a \lor b)$ (here $\square$ means globally) would be converted to $\LTLGlobally{p0}$, where $p0$ is boolean predicate representing $a \lor b$\@.
% As part of this process, we also create functions from model state (a python dictionary containing the current values of the model) to predicate values\@.
\NewParagraph{Monitor}
We provide a quick refresher on $\BuchiAutomaton$\@.
For details, see Subsection~\ref{\LabelSubsection{\CurrentChapter}{Preliminaries}{Buchi Automata}}\@.
A $\BuchiAutomaton$ is a nondeterministic automaton with transition guards\@.
Thus from a given state, $\BuchiAutomaton$ can transition to any other state provided a transition to that state exists and the associated guard condition is true\@.
Within the formal definition, these guards are encoded into the transition function\@.
Some of the states in the $\BuchiAutomaton$ are `accepting' states\@.
The $\BuchiAutomaton$ accepts a trace if there exists a sequence where the $\BuchiAutomaton$ is infinitely often in an accepting state\@.
\par{}
To mimic this behavior, the monitor takes as input a set of states that the $\BuchiAutomaton$ could be in along with the current model state\@.
The current model state provides all the necessary information to determine if a transition guard is true\@.
This, combined with the possible states, is used to create a new set of possible states\@.
If there are no possible states, then there is no longer any way for the specification to be true, meaning it must be false\@.
If there is a possible state that is an accepting state with a transition to itself and the guard is always true, then the specification is guaranteed to be true\@.
Otherwise, the specification could still prove to be true or false (unknown)\@.
The monitor returns both the new set of possible states and the verdict to the user\@.
If a violation occurs, the monitor can be `reset'\@.
This is important as we want our monitor to be repeatably usable; without a reset it would have to continuously report that a violation occurred\@.
\NewSubsection{nuXmv Model Generation}
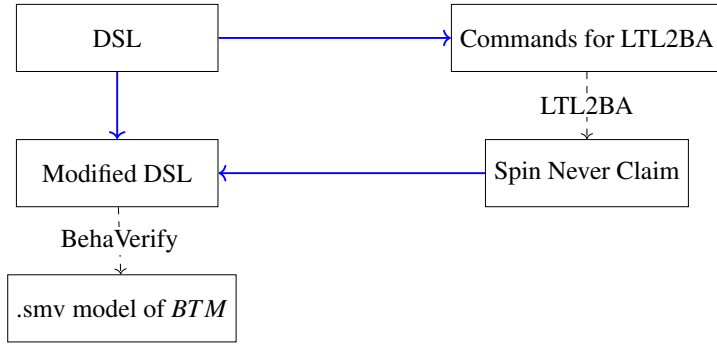
\begin{figure}
  \begin{minipage}{.6\linewidth}
    \begin{adjustbox}{max width = \linewidth}
      \begin{tikzpicture}[
    box/.style = {draw, minimum width=3cm, minimum height=1cm, align=center},
    arrow/.style = {->, thick},
    label/.style = {midway, fill=white, inner sep=1pt}
]

% Nodes
\node[box] (dsl) {DSL};
\node[box, below=of dsl] (m_dsl) {Modified DSL};
\node[box, right=of dsl, xshift=70pt] (ltl2ba_cmd) {Commands for LTL2BA};
\node[box, below=of ltl2ba_cmd] (spin_mon) {Spin Never Claim};
\node[box, below=of m_dsl, xshift=2pt] (smv) {.smv model of $\BehaviorTreeMonitor$};

% Arrows
\draw[arrow, blue] (dsl) -- (m_dsl);
\draw[arrow, blue] (dsl) -- (ltl2ba_cmd);
\draw[->, dashed] (ltl2ba_cmd) -- node[label] {LTL2BA} (spin_mon);
\draw[arrow, blue] (spin_mon) -- (m_dsl);
\draw[->, dashed] (m_dsl) -- node[label] {BehaVerify} (smv);

\end{tikzpicture}
    \end{adjustbox}
  \end{minipage}
  \begin{minipage}{.39\linewidth}
    \caption{
      Diagram of how BehaVerify generates {.}smv files for $\BehaviorTree$s with monitors that can be verified using nuXmv\@.
      LTL2BA is a tool for converting an LTL specification to a $\BuchiAutomaton$\@.
      Spin is a model checker, and a Never Claim can be checked using Spin\@.
      Solid blue arrows mark new contributions\@.
    }\label{\LabelObjectHere{SMV Pipeline}}
  \end{minipage}
\end{figure}
The nuXmv pipeline is very similar to the Python pipeline, and as such we will avoid going into the inner workings of this pipeline\@.
The main difference is that we create a monitor using the DSL of BehaVerify allowing BehaVerify to create a {.}smv model for use with nuXmv\@.

\NewSection{Monitor Comparison}
We created two scaling scenarios and ran timing comparisons for the generated monitors created by BehaVerify and Copilot\@.
We planned to compare two monitors generated by NuRV, but did not ultimately do so (see Subsection~\ref{\LabelSubsection{\CurrentChapter}{\CurrentSection}{NuRV}} for details)\@.
Additionally, we utilized the generated Python code with monitors to generate example traces\@.
While these example traces are not conclusive proof, they were sufficient to demonstrate some differences in the generated monitors\@.
The results demonstrate that the monitors generated by BehaVerify are not outclassed by existing tools and in some cases are preferable from a correctness standpoint.
Furthermore, this demonstrates the versatility of our setup; it is fairly painless to bring in outside monitors should the need arise\@.
\par{}
The rest of this section will describe the scaling scenarios, present the results, and then reason about the results\@.
\NewSubsection{Scenarios}
For our scenarios, a drone navigates a grid and tries to reach a destination\@.
Once the destination is reached, a new destination is randomly generated\@.
We generated grids from size 10 by 10 to 50 by 50 in two styles: dense fixed and sparse random\@.
The dense fixed grids start with a 5 by 5 grid and copies this layout to fill the entire grid\@.
The sparse random grids are randomly generated\@.
See Figure~\ref{\LabelObjectHere{Grids}} for details\@.
\begin{figure}%[t]
  \begin{minipage}{.48\linewidth}
    \includegraphics[width=\linewidth,keepaspectratio]{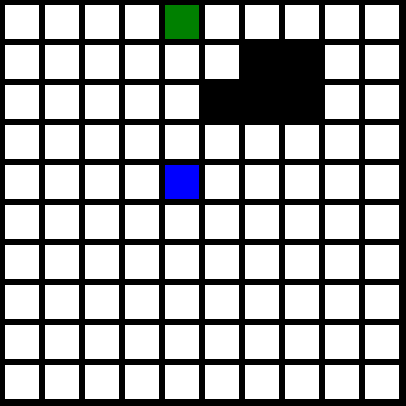}
  \end{minipage}
  \begin{minipage}{.48\linewidth}
    \includegraphics[width=\linewidth,keepaspectratio]{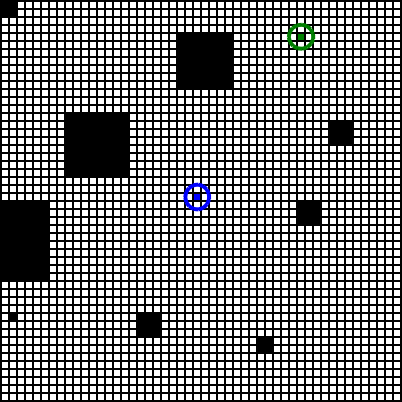}
  \end{minipage}\\
  \begin{minipage}{.48\linewidth}
    \includegraphics[width=\linewidth,keepaspectratio]{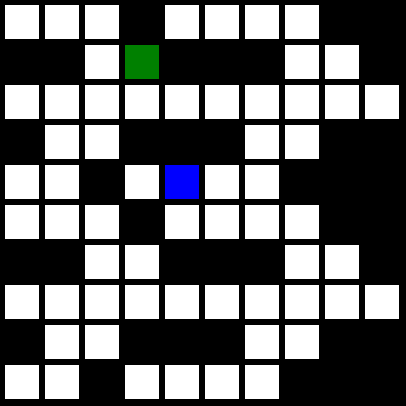}
  \end{minipage}
  \begin{minipage}{.48\linewidth}
    \includegraphics[width=\linewidth,keepaspectratio]{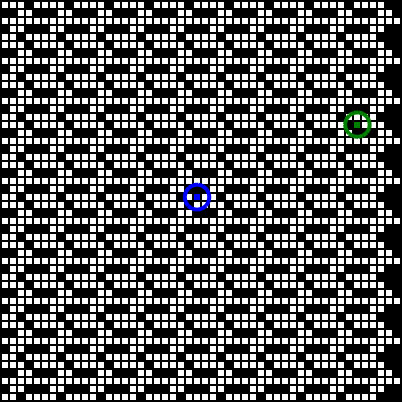}
  \end{minipage}
  \caption{
    Images representing some of the grids used for the scaling experiments\@.
    The upper grids are sparse and were randomly generated\@.
    The lower grids are dense and were created by copying a 5 by 5 grid repeatedly\@.
    The left grids are 10 by 10 and the right grids are 50 by 50\@.
    Black squares are obstacles, the blue square is the drone, and the green square is the destination\@.
  }\label{\LabelObjectHere{Grids}}
\end{figure}
\par{}
At each time step, the drone attempts to move towards the target according to the control logic\@.
We want the drone to satisfy the following specifications
\begin{equation*}
  \begin{split}
    \LTLSafety &= \LTLGlobally(\forall (x_o, y_o) \in Obs, (x_d, y_d) \neq (x_o, y_o))\\
    \LTLLiveness &= \LTLGlobally(\LTLFinally((x_d, y_d) = (x_g, y_g) \lor \exists (x_o, y_o) \in Obs \SuchThat (x_g, y_g) = (x_o, y_o)))
  \end{split}
\end{equation*}
Here $Obs$ is a predetermined finite set of obstacles\@.
$\LTLSafety$ is a safety specification that states that the drone's position $(x_d, y_d)$ is never equal to the position of an obstacle $(x_o, y_o)$\@.
$\LTLLiveness$ is a liveness specification that states that the drone's position is eventually equal to the destination $(x_g, y_g)$, or the destination is an obstacle\@.
We utilize the quantifiers $\forall, \exists$ (for all and exists, respectively) for convenience here; in practice we write out each individual obstacle\@.
%Finally, $(a, b) = (c, d) \iff (a = c) \land (b = d)$\@.
\par{}
Then the specifications we monitor are
\begin{equation*}
  \begin{split}
    \LTLSafetyA &= \LTLGlobally(\forall (x_o, y_o) \in Obs, (x_d + (x_{\Delta} * s), y_d + (y_{\Delta} * s)) \neq (x_o, y_o))\\
    \LTLLivenessA &= \LTLGlobally
    \begin{pmatrix*}
      (x_{\Delta}, y_{\Delta}) = (1, 0) \implies \LTLNext((x_{\Delta}, y_{\Delta}) \neq (-1, 0)) \land\\
      (x_{\Delta}, y_{\Delta}) = (-1, 0) \implies \LTLNext((x_{\Delta}, y_{\Delta}) \neq (1, 0)) \land\\
      (x_{\Delta}, y_{\Delta}) = (0, 1) \implies \LTLNext((x_{\Delta}, y_{\Delta}) \neq (0, -1)) \land\\
      (x_{\Delta}, y_{\Delta}) = (0, -1) \implies \LTLNext((x_{\Delta}, y_{\Delta}) \neq (0, 1))
    \end{pmatrix*}
  \end{split}
\end{equation*}
$(x_d, y_d)$ is the current location of the drone, $(x_{\Delta}, y_{\Delta})$ is one of $(1, 0)$, $(-1, 0)$, $(0, 1)$, $(0, -1)$, $(0, 0)$, describing the possible directions the drone moves in, and $s$ is the speed of the drone (either 1 or 2)\@.
Thus the safety specification monitor is violated if we are on a collision course and the liveness specification monitor is violated if we do a 180 turn\@.
We note that both specifications being monitored are safety specifications; however, we will refer to the second as a liveness specification as it is being used to ensure the liveness specification is not violated\@.
Please note the following: we are \textit{not} claiming that by monitoring these conditions any $\BehaviorTree$ will satisfy the desired specifications; rather, the purpose is to demonstrate that the monitors can be used to correct specific flaws in a $\BehaviorTree$\@.
Furthermore, it is possible to design and insert these monitors into the $\BehaviorTree$ without the use of a special tool; however, such a task may prove more complex then writing an LTL specification and utilizing our tool\@.
\par{}
By default, the drone will try to move 2 squares; if either the safety monitor or liveness monitor are triggered, it will only move one square\@.
The safety monitor ensures that the drone does not move into obstacles\@.
The liveness monitor ensures that the drone escapes potential loops, as seen in Figure~\ref{\LabelObjectHere{Loop}}\@.
Thus both monitors are necessary for the drone to function as intended\@.
Note that this example is not meant to illustrate good programming practice for drone controllers; there are no doubt better methods by which to control a drone\@.
Rather, the purpose of this example is to demonstrate how one can use the monitors to ensure a system functions correctly\@.
\begin{figure}
  \begin{minipage}{.32\linewidth}
    \includegraphics[width=\linewidth,keepaspectratio]{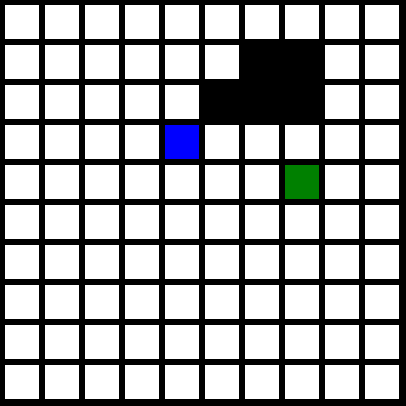}
  \end{minipage}
  \begin{minipage}{.32\linewidth}
    \includegraphics[width=\linewidth,keepaspectratio]{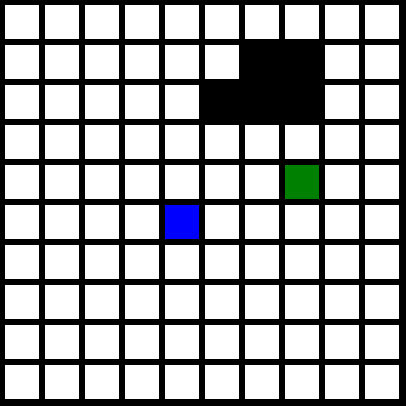}
  \end{minipage}
  \begin{minipage}{.32\linewidth}
    \includegraphics[width=\linewidth,keepaspectratio]{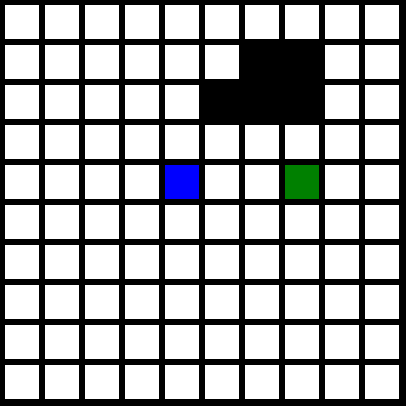}
  \end{minipage}
  \caption{
    Pictures are ordered left to right\@.
    The drone (blue) is trying to reach the destination (green) while avoiding obstacles (black)\@.
    In the left image, the control logic tells the drone to go down\@.
    Because neither monitor reported a violation, the drone moved 2 squares and is now in the situation shown in the middle image\@.
    The control logic now tells the drone to go up\@.
    The liveness monitor reports a violation; if we go up two squares, the drone will be stuck in a loop\@.
    Therefore, the drone only goes up one square and is now in the situation shown in the right image\@.
    If the drone has not been equipped with the liveness monitor, it would have gone back to the state shown in the left image, then middle, then left, etc\@.
  }\label{\LabelObjectHere{Loop}}
\end{figure}
\NewSubsection{Motivation}
We were originally creating a controller for a drone in virtual neighborhood simulated using AirSim (see Figure~\ref{\LabelObjectHere{AirSim}})\@.
The drone would fly at a fixed height and knew where obstacles were before hand\@.
We created a grid-world abstraction of the problem and created a controller that we were able to verify navigated correctly under certain assumptions\@.
One of the assumptions was that the drone would always move at most one tile at a time\@.
While it is simple to ensure that this is the case, it requires flying the drone slowly at all times, which is not desirable\@.
Flying the drone at faster speeds, however, created both safety and liveness violations\@.
Thus by equipping our $\BehaviorTree$ with the described monitors, we were able to safely increase speed without compromising safety\@.
Finally, we were able to utilize nuXmv to verify that the $\BehaviorTreeMonitor$ is safe (see Section~\ref{\LabelSection{\CurrentChapter}{Design Time Verification}} for details)\@.
\begin{figure}
  \includegraphics[width=\linewidth,keepaspectratio]{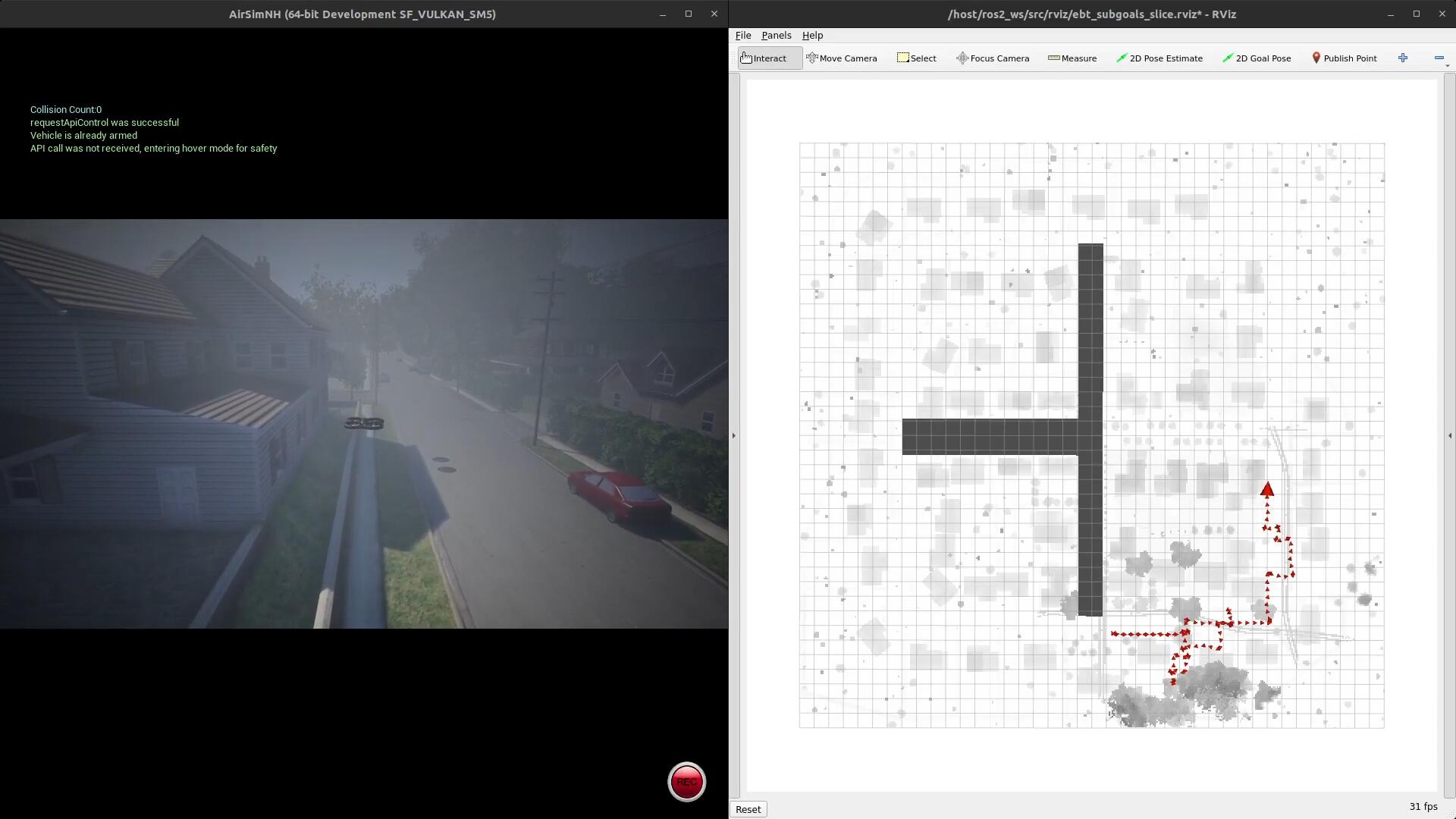}
  \caption{
    A screenshot of the drone flying in AirSim and a grid visualization\@.
  }\label{\LabelObjectHere{AirSim}}
\end{figure}
\NewSubsection{Results}
\begin{figure}
  \begin{minipage}{.48\linewidth}
    \includegraphics[width=\linewidth,keepaspectratio]{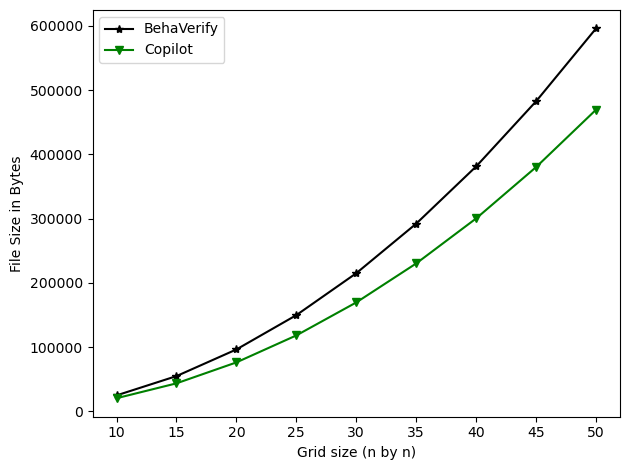}
  \end{minipage}
  \begin{minipage}{.48\linewidth}
    \includegraphics[width=\linewidth,keepaspectratio]{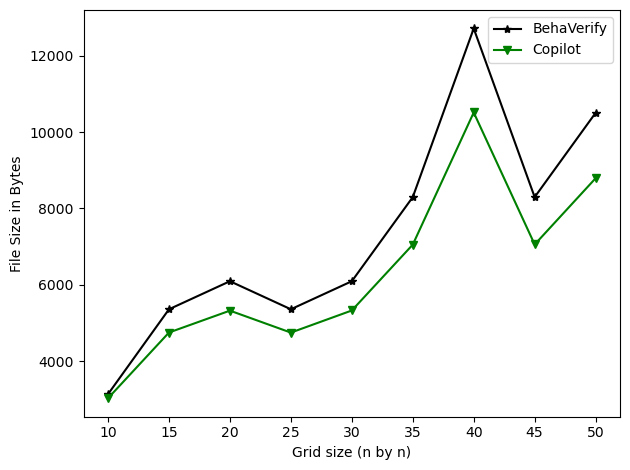}
  \end{minipage}\\
  \begin{minipage}{.48\linewidth}
    \includegraphics[width=\linewidth,keepaspectratio]{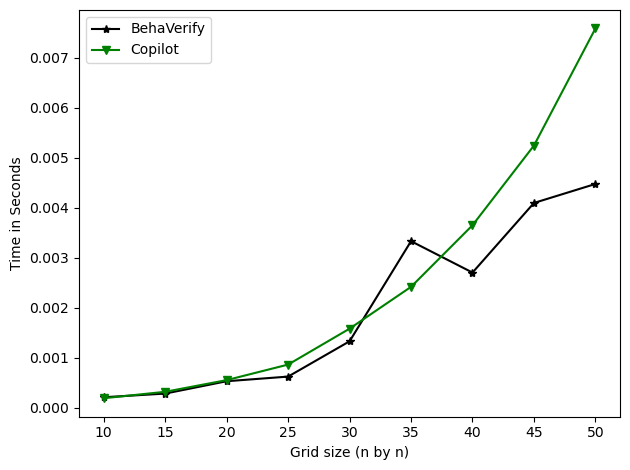}
  \end{minipage}
  \begin{minipage}{.48\linewidth}
    \includegraphics[width=\linewidth,keepaspectratio]{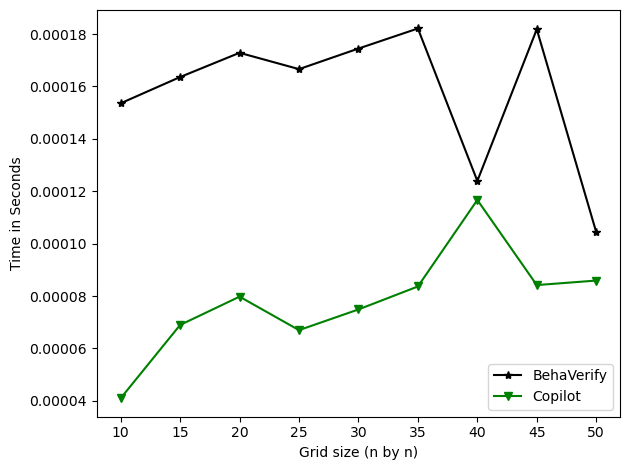}
  \end{minipage}
  \caption{
    Top left: file sizes of safety monitors for dense fixed\@.
    Top right: file sizes of safety monitors for sparse random\@.
    Bottom left: median time (in seconds) for safety and liveness monitors on dense fixed\@.
    Bottom right: median time (in seconds) for safety and liveness monitors on sparse random\@. 
    We did not include the size of liveness monitors as it did not change with the size of the grid\@.
    The timing results are the median of 10000 runs\@.
    Each run had 1000 iterations (number of times the drone tried to move)\@.
    We subtracted how long it took to run the code without a monitor\@.
  }\label{\LabelObjectHere{Comparison}}
\end{figure}
All results (Figure~\ref{\LabelObjectHere{Comparison}}) were generated on a computer with a 24 core 13th Gen Intel(R) Core(TM) i7{-}13700K and 64GB of DDR5 RAM\@.
Results were run in a quiet environment when no other user run processes were active\@.
The code for the experiments is available at~\footnote{\url{https://github.com/verivital/behaverify/tree/main/REPRODUCIBILITY/2024_FMAS_BTM}}\@.
\NewSubsection{Analysis}
%We divide the analysis into three parts: Timing, File Size, and Correctness\@.
\NewParagraph{Timing}
We ran $10000$ simulation runs with $1000$ iterations (number of times the drone tried to move) and took the median for both Copilot and BehaVerify\@.
To ensure that we are comparing only the monitors, we also ran a version of the code with no monitor with the same settings\@.
This monitorless value was then subtracted out from the timing results for both BehaVerify and Copilot\@.
We believe that the timing results demonstrate that the monitors generated by BehaVerify and Copilot are reasonably close\@.
As expected, the dense fixed grid pattern produces a far more linear timing relation than the sparse random grid\@.
This is because the number of obstacles in the dense fixed grid is always greatly increasing, while the number of obstacles in the sparse random setup is random, and it is the number of obstacles that is largely responsible for the complexity of monitoring the specifications\@.
\NewParagraph{File Size}
We measured the file size as an indicator of the complexity of the monitoring algorithm\@.
While BehaVerify clearly generates larger files than Copilot, we do not believe that the difference is sufficient to prefer Copilot's Monitors\@.
This is in contrast to NuRV, which is discussed below in Subsection~\ref{\LabelSubsection{\CurrentChapter}{\CurrentSection}{NuRV}}\@.
\NewParagraph{Correctness}
The safety monitors that Copilot generated worked as intended, but the liveness monitor had an issue: it would report a violation with a one-step delay\@.
Specifically, if a violation occurred in the $i^{th}$ state, then Copilot would report it in the $i+1^{th}$ step\@.
This behavior is documented in the Copilot tutorial in example 7 \footnote{\url{https://copilot-language.github.io/downloads/copilot_tutorial.pdf}}\@.
By contrast, both the safety and liveness monitors generated by BehaVerify worked as intended, ensuring the drone functioned as intended\@.
\par{}
This brings us to an important note about how our process is laid out\@.
While BehaVerify is capable of generating monitors, the generated Python code can utilize \textit{any} provided monitor\@.
Indeed, we confirmed that it is possible to utilize Copilot for the safety monitor and BehaVerify for the liveness monitor and that this works as intended\@.
\NewSubsection{NuRV}
Because BehaVerify creates {.}smv files, we originally tried using the output of BehaVerify as input to NuRV\@.
Unfortunately, a variety of issues prevented this from being feasible\@.
For instance, NuRV monitors are aware of the transition system\@.
This enable NuRV monitors to potentially detect violations well in advance or to verify liveness conditions, but it also means the files are much larger\@.
To combat this, we created simplified {.}smv models with much simpler transition systems\@.
This proved ineffective; the smallest file generated by NuRV was 26.04 MB, while the largest file generated by BehaVerify was 582 KB\@.

\NewSection{Design Time Verification}
BehaVerify was originally created for Design Time Verification\@.
As such, when approaching the topic of Runtime Verification, we were interested if we could use Design Time Verification for the Runtime Monitors\@.
As such, we translated the monitors that were created by BehaVerify back into the DSL for BehaVerify and then utilized nuXmv to verify that the $\BehaviorTree$ with monitors satisfied both the safety and liveness specification\@.
While we created such translations for each combination of grid type and grid size presented in Section~\ref{\LabelSection{\CurrentChapter}{Monitor Comparison}}, some of the resulting models proved to complex for liveness analysis in nuXmv\@.
The results for safety verification can be seen in Figure~\ref{\LabelObjectHere{Design Times}}\@.
As you can see, it is entirely feasible to use design time verification to confirm that the safety monitors are correct and ensure the system works as intended\@.
If the safety monitor is removed, nuXmv will demonstrate that the system is not safe by providing a counter example trace resulting in a crash\@.
If the liveness monitor is removed, nuXmv will demonstrate that the system can get stuck in a loop by providing a counter example trace\@.
\begin{figure}
  \begin{minipage}{.48\linewidth}
    \includegraphics[width=\linewidth,keepaspectratio]{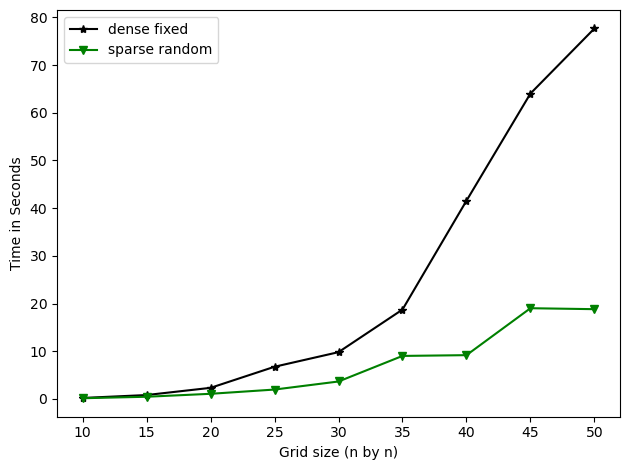}
  \end{minipage}
  \begin{minipage}{.50\linewidth}
    \caption{
      The graph shows the time (in seconds) to verify that the $\BehaviorTreeMonitor$ is safe (does not crash into obstacles)\@.
      The verification was done with nuXmv\@.
      Liveness specifications are considerably harder to verify, with the 9 by 9 sparse grid taking about 25 minutes to verify\@.
      As this is the easiest grid we have for this task, we did not complete liveness verification for any other grids\@.
    }\label{\LabelObjectHere{Design Times}}
  \end{minipage}
\end{figure}
\par{}
The liveness situation is somewhat trickier, as the specifications are substantially harder to verify\@.
However, considering the fact that the same liveness monitor is used for all grids and that we verified it for one grid, even this limited verification process can provide some evidence to indicate that the monitor is correct\@.
As with the safety monitor, the removal of the liveness monitor results in a specification violation that nuXmv detects\@.
In this case, nuXmv returns a counterexample where the drone becomes stuck in a loop, going back and forth between two points without reaching the destination\@.

\NewSection{Conclusions and Future Work}
We presented a formal problem statement for incorporating contingency monitors within $\BehaviorTree$s, thus creating $\BehaviorTreeMonitor$s\@.
On the implementation side, we expanded the DSL of BehaVerify to incorporate these monitors, and demonstrated that our code is capable of generating implementations of the monitors that are on par with existing tools\@.
However, our overall approach also brings the advantage of Design Time Verification for the entire $\BehaviorTreeMonitor$\@.
We subsequently hope to expand the target range of BehaVerify, specifically to create {.}cpp implementations that make use of BehaviorTrees.cpp\@.

\section*{Acknowledgements}
The material presented in this paper is based upon work supported by the National Science Foundation (NSF) through grant numbers 2220426 and 2220401, the Defense Advanced Research Projects Agency (DARPA) under contract number FA8750-23-C-0518, and the Air Force Office of Scientific Research (AFOSR) under contract numbers FA9550-22-1-0019 and FA9550-23-1-0135. Any opinions, findings, and conclusions or recommendations expressed in this paper are those of the authors and do not necessarily reflect the views of AFOSR, DARPA, or NSF.

\bibliographystyle{eptcs}
\bibliography{./Utility/Bibliography}

\end{document}